\documentclass[preprint,12pt,authoryear]{elsarticle}

\usepackage{amssymb}
\usepackage{amsmath}
\usepackage{url}
\usepackage{bbm}
\usepackage{booktabs}
\usepackage[table]{xcolor}

\journal{Reliability Engineering \& System Safety}

\begin{document}

\begin{frontmatter}



\title{Predictive Multiplicity in Survival Models: A Method for Quantifying Model Uncertainty in Predictive Maintenance Applications} 


\author[label1]{Mustafa Cavus} 

\affiliation[label1]{organization={Eskisehir Technical University, Department of Statistics},
            city={Eskisehir},
            postcode={26555}, 
            country={Turkiye}}

\begin{abstract}
In many applications, especially those involving prediction, models may yield near-optimal performance yet significantly disagree on individual-level outcomes. This phenomenon, known as \textit{predictive multiplicity}---formally defined in binary, probabilistic, and multi-target classification---undermines the reliability of predictive systems. However, its implications remain unexplored in the context of \textit{survival analysis}, which involves estimating the time until a \textit{failure} or similar event, while properly handling censored data. We frame predictive multiplicity as a critical concern in \textit{survival-based models} and introduce formal measures—\textit{ambiguity}, \textit{discrepancy}, and \textit{obscurity}—to quantify it. This is particularly relevant for downstream tasks such as \textit{maintenance scheduling}, where precise individual risk estimates are essential. Understanding and reporting predictive multiplicity helps \textit{build trust} in models deployed in high-stakes environments. We apply our methodology to benchmark datasets from \textit{predictive maintenance}, extending the notion of multiplicity to survival models. Our findings show that \textit{ambiguity} steadily increases, reaching up to $40$--$45\%$ of observations; \textit{discrepancy} is lower but exhibits a similar trend; and \textit{obscurity} remains mild and concentrated in a few models. These results demonstrate that multiple accurate survival models may yield conflicting estimations of \textit{failure risk} and \textit{degradation progression} for the same equipment. This highlights the need to explicitly measure and communicate predictive multiplicity to ensure reliable decision-making in \textit{process health management}.

\end{abstract}




\end{frontmatter}

\section{Introduction}

In today's highly competitive industrial landscape, even brief periods of downtime can significantly hinder operational efficiency, often reducing productivity by as much as $5\%$ to $20\%$ across a wide range of sectors \citep{kane_et_al_2022}, highlighting the critical role of effective \textit{process health management} in minimizing disruptions and maintaining performance. Traditional maintenance strategies have gradually evolved—from reactive maintenance, where action is taken only after a \textit{failure} occurs, to preventive maintenance, which involves scheduled interventions regardless of equipment condition, and condition-based maintenance, which relies on real-time monitoring to trigger maintenance activities—laying the groundwork for the emergence of predictive maintenance as a proactive and efficient solution. Predictive maintenance refers to a data-driven approach to anticipating \textit{equipment degradation} and predicting \textit{failure} before they occur, enabling timely interventions that reduce unplanned downtime and optimize maintenance costs. Building on this evolution, predictive maintenance leverages various machine learning approaches tailored to different prediction needs and data characteristics.

In predictive maintenance, regression models are used to predict continuous variables, such as estimating how much longer a pump will function before failure, while classification models help categorize equipment into failure-risk groups, for instance, classifying machinery as high-risk or low-risk for \textit{failure}. Additionally, clustering algorithms group similar operational conditions to detect abnormal patterns, like identifying clusters of vibration data that signal early signs of wear, and survival models (also known as event history, duration, or reliability models) estimate the time until a failure event occurs, offering a more precise prediction of failure timing and capturing degradation progression over time. Unlike the other approaches, survival models provide a distinct advantage of modeling the time-to-failure, which allows for more effective maintenance scheduling, as it predicts not just whether a \textit{failure} will occur, but when it is likely to occur. Survival models are particularly valuable due to their natural ability to handle commonly encountered censored data—observations where a failure has not yet occurred—making them better suited for modeling the probabilistic evolution of \textit{process health management} over time \citep{lillelund_et_al_2024,yang_et_al_2022}.

Although modeling processes can make significant contributions to solving such industry-specific problems, developing an optimal predictive model often involves a highly challenging process. A typical strategy in the modeling process involves training models under multiple configurations and choosing the one that yields the highest prediction performance. However, this approach prompts a critical issue: identifying the best model becomes challenging when multiple candidates demonstrate near-optimal performance. This challenge is called \textit{Rashomon effect}, which refers to the existence of many near-optimal models for a task \citep{breiman_2001}. It becomes exponentially harder in the case of \textit{predictive multiplicity}, which is the ability of near-optimal models to assign conflicting predictions to observations for the same prediction task \citep{marx_et_al_2020}. This concept enables stakeholders to make more informed choices in model selection, validation, and \textit{model uncertainty} quantification \citep{ali_et_al_2021, sokol_et_al_2024} by developing predictive multiplicity metrics to measure the variation between predictions of competing models. In this context, model uncertainty refers to the uncertainty arising from the lack of knowledge about the true underlying model and is often referred to as \textit{epistemic uncertainty}. This type of uncertainty can be reduced by acquiring more data or improving the model, and it is distinct from \textit{aleatoric uncertainty} (also known as \textit{data uncertainty}), which arises from inherent variability \citep{huellermeier_et_al_2021, abdar_et_al_2021, gruber_et_al_2023}. Predictive multiplicity appears first and the relevant metrics are proposed in the context of binary classification tasks \citep{marx_et_al_2020}, then probabilistic classification \citep{watson_et_al_2023}, and multi-target classification \citep{watson_et_al_2023b}.

Similarly, predictive multiplicity in predictive maintenance leads to diverse model predictions, which may introduce significant uncertainty in decision making \citep{rudin_et_al_2024}, especially when multiple models suggest conflicting maintenance actions. This inconsistency not only complicates the interpretation of \textit{process health management} but also increases the likelihood of mispredicting the severity of equipment issues, leading to either premature maintenance or costly unplanned downtimes, thus undermining both cost-effectiveness and system reliability \citep{du_et_al_2024}. Moreover, such conflicting outputs may erode trust in predictive systems, making engineers and decision-makers reluctant to rely on them in critical planning contexts. Integrating model uncertainty further enhances trust in these models, addressing key challenges in reliability and decision-making \citep{iversen_et_al_2023}. However, it is important to note that the predictive multiplicity metrics developed for various prediction tasks do not yet address the survival task.

This paper responds to the identified shortcoming by studying the phenomenon of predictive multiplicity in survival-based predictive maintenance settings, proposing formal metrics to quantify it, and demonstrating its implications for risk estimation under uncertainty. We demonstrate this through a case study based on a survival task in the aerospace domain, focusing on predicting aircraft engine failure risks using the Random Survival Forests model \citep{ishwaran_et_al_2008} trained on the CMAPSS dataset—a task of growing importance in predictive maintenance and reliability engineering. To the best of our knowledge, this is the first paper to systematically examine multiplicity in the survival models, addressing a critical gap in model reliability and decision robustness. Moreover, by calculating the degree of predictive multiplicity, i.e., how many different predictions are possible, we can not only gain insights into the model uncertainty \citep{rudin_et_al_2024} but also, by using multiple models \citep{li_et_al_2024, biecek_et_al_2024}, obtain different perspectives on the task at hand, a practice that has gained popularity in recent years. Specifically, we make the following contributions:

\begin{enumerate}
    \item We extend the concept of predictive multiplicity to survival models commonly used in predictive maintenance, enhancing \textit{process health management} through a better understanding of \textit{equipment degradation} and \textit{failure} risk.
    \item We introduce three tailored metrics—\textit{ambiguity}, \textit{discrepancy}, and \textit{obscurity}—that quantify the extent of disagreement among equally well-performing survival models, helping in \textit{validating and building confidence} in the model's predictive capabilities.
    \item We apply our methodology to real-world predictive maintenance data using the CMAPSS dataset. Our empirical findings demonstrate that survival models may yield significantly different predictions for the same equipment even under similar performance metrics, raising critical concerns for \textit{maintenance decision-making} and \textit{failure} risk estimation.
\end{enumerate}

The remainder of this paper is organized as follows: Section 2 introduces the fundamental concepts of predictive multiplicity for survival models, emphasizing its role in improving \textit{process health management}. Section 3 details the experimental setup, while Section 4 summarizes the results with a focus on \textit{failure} prediction accuracy and \textit{equipment degradation}. Finally, Section 5 presents the concluding remarks and key takeaways, highlighting the importance of \textit{validating and building confidence} in survival models for \textit{maintenance decision-making}.

\section{Related Work}

\subsection{Predictive maintenance}

The research on predictive maintenance (PM) has evolved significantly over the past two decades, driven by advances in sensor technology, the increasing availability of data, and growing computational capabilities. Early PM systems were primarily rule-based or depended on physical degradation models \citep{jardine_et_al_2006}. These approaches demanded extensive domain expertise and predefined failure mechanisms, which limited their scalability and flexibility across different equipment types.

A major shift occurred in the 2010s with the widespread adoption of machine learning techniques, coinciding with the digitization of industrial systems. Supervised models such as decision trees, support vector machines, and neural networks began to be used for classifying or predicting failures using labeled historical data \citep{schwabacher_2005, heng_et_al_2009}. During this period, research focused on improving predictive accuracy through feature engineering, signal processing, and model tuning.

As the field matured, attention turned from algorithmic improvements to more systemic challenges. One critical issue was the class imbalance inherent in failure data, prompting the use of resampling strategies and cost-sensitive learning \citep{sipos_et_al_2014, zhao_et_al_2019}. Another was the limited adaptability of static models in dynamic production environments, which spurred interest in concept drift detection and adaptive learning methods \citep{widodo_et_al_2007}.

More recently, the predictive maintenance landscape has been shaped by the principles of Industry 4.0, which emphasizes cyber-physical systems, real-time analytics, and data-driven decision-making. As \citet{pinciroli_et_al_2023} highlights, Industry 4.0 facilitates the integration of vast, heterogeneous data, such as sensor signals, images, and operational logs, enabling more effective anomaly detection, diagnostics, and failure prediction through advanced AI algorithms. This transition has reduced reliance on subjective expert knowledge and enabled the design of PM systems that dynamically adapt to system conditions. Accordingly, deep learning models like LSTMs and autoencoders have been increasingly used for time series forecasting in high-dimensional settings \citep{zhao_et_al_2019, babu_et_al_2016}.

Furthermore, hybrid frameworks that combine physics-based reasoning with data-driven models have gained prominence for their enhanced generalizability across domains \citep{shoorkand_et_al_2024}. Recent reviews have underscored the need to situate PM approaches within the broader context of smart manufacturing, highlighting the strategic role of maintenance in enhancing sustainability, resilience, and system-wide optimization \citep{bousdekis_et_al_2020}.

\subsection{Survival predictive maintenance}

Survival models have become a crucial tool in PM due to their natural ability to handle time-to-event data, which is essential for estimating RUL and failure risk of machinery \citep{holmer_et_al_2023}. Unlike traditional classification or regression models that focus on whether a failure will occur, survival models estimate when it is likely to happen. However, the number of studies explicitly applying survival analysis within PM remains relatively limited. Among the notable contributions, \citet{alabdallah_et_al_2024} notes that these models yield survival or hazard functions, providing time-dependent probabilities of failure, making them especially relevant for PM scenarios with uncertain degradation dynamics.

A key strength of survival models is their capacity to deal with censored data—a common feature in real-world PM datasets where not all failures are observed during the study period. \citet{zeng_et_al_2023} explains that survival analysis explicitly accounts for right-censoring, improving the robustness of predictions by incorporating the uncertainty of unobserved events. Similarly, \citet{yang_et_al_2022} emphasizes the utility of survival analysis in mobile equipment maintenance, where incomplete and irregular logging is typical.

Recent advances have expanded survival modeling beyond classical methods. For instance, \citet{ishwaran_et_al_2008} introduces random survival forests—a non-parametric alternative that can capture complex survival curves without assuming proportional hazards. This flexibility is particularly useful for modeling diverse equipment behaviors under varying operational conditions. Extending this further, \citet{rahat_et_al_2023} proposes a framework to reformulate run-to-failure datasets for survival analysis, demonstrating a seamless integration between traditional RUL prediction and survival-based approaches. 

\subsection{Uncertainty quantification in predictive maintenance}

Quantifying model uncertainty has emerged as a critical aspect of trustworthy and reliable PM systems, particularly in applications where erroneous maintenance decisions can lead to substantial operational and financial losses \citep{kane_et_al_2022, guillaume_et_al_2020}. While traditional models often focus solely on predictive accuracy, recent work highlights the importance of incorporating both aleatoric and epistemic uncertainties to assess model confidence better and facilitate informed decision-making \citep{he_et_al_2023, iversen_et_al_2023}.

Importantly, \citet{lillelund_et_al_2024} provides probabilistic RUL estimates, making them a strong candidate for uncertainty-aware maintenance strategies. Bayesian approaches, in particular, enable explicit modeling of uncertainty, allowing for individualized confidence estimates \citep{zeng_et_al_2023}.

However, the growing complexity of predictive models has introduced new challenges. The phenomenon of predictive multiplicity—where multiple, equally well-performing models make diverging predictions—has been recognized as a key contributor to model uncertainty and mistrust \citep{rudin_et_al_2024, du_et_al_2024}. This is especially concerning in PM contexts, where model outputs can trigger maintenance actions with real-world consequences. \citet{ali_et_al_2021} and \citet{sokol_et_al_2024} argue that such multiplicity reflects epistemic uncertainty and may signal underspecification of the model, aligning with the broader Rashomon effect observed in complex data-driven systems.

Recent work by \citet{yardimci_et_al_2025} introduces a novel approach to uncertainty quantification in predictive maintenance by leveraging the Rashomon perspective. Instead of relying on a single best-performing model, their method constructs a Rashomon survival curve that captures the prediction range of multiple, similarly accurate survival models. This approach explicitly visualizes epistemic uncertainty stemming from predictive multiplicity and demonstrates how model agreement and variability evolve. Their findings on the CMAPSS dataset reveal that censoring time significantly affects prediction uncertainty, underscoring the risks of single-model reliance in high-stakes PM scenarios. Unlike traditional ensemble methods, this strategy emphasizes interpretability and robustness by preserving model diversity, offering a practical solution to underspecification in survival-based RUL estimation.

Recent studies propose various strategies to mitigate this uncertainty. Some advocate for ensemble or multiverse modeling to reveal the sensitivity of predictions to design choices \citep{simson_et_al_2024}, while others explore explainability-driven solutions to expose uncertainty sources at both data and model levels \citep{kargar_et_al_2024, baniecki_et_al_2025}. 

Despite this growing body of work, a commonly accepted framework for uncertainty quantification in remains elusive \citep{gruber_et_al_2023}. As such, there is a pressing need for models that can not only predict failures, but also communicate the confidence in these predictions in a transparent and interpretable manner—paving the way for more resilient and actionable PM systems.

\section{Methodology}
\subsection{Preliminaries}

We consider a survival task on a dataset $\mathcal{D} = \{(\mathbf{x}_i, t_i, \eta_i)\}_{i=1}^n$ of $n$ observations. Each example consists of a feature vector $\mathbf{x}_i = [1, x_{i1}, \ldots, x_{id}] \in X \subseteq \mathbb{R}^{d+1}$, an event or censoring time $t_i \in \mathbb{R}_{\geq 0}$, and an event indicator $\eta_i \in \{0, 1\}$, where $\eta_i = 1$ indicates that the event of interest (e.g., system breakdown, engine failure, etc.) occurred at time $t_i$, and $\eta_i = 0$ indicates that the observation was right-censored at time $t_i$. With this dataset, we train a survival model $f: X \times \mathbb{R}_{\geq 0} \rightarrow [0, 1]$ that estimates the risk of event occurrence up to time $t$ for a given input $\mathbf{x}_i$:  
\begin{equation}
    f(\mathbf{x}_i, t) := \Pr(T \leq t \mid \mathbf{x}_i),
\end{equation}

\noindent where $T$ is a non-negative random variable representing the time to event. This function corresponds to the conditional cumulative distribution function (CDF) of the event time, capturing the cumulative incidence or risk over time.

For completeness, we recall that the conditional \textit{survival function} is defined as:
\begin{equation}
    S(t \mid \mathbf{x}_i) := \Pr(T > t \mid \mathbf{x}_i) = 1 - f(\mathbf{x}_i, t),
\end{equation}
and describes the probability that the event has not yet occurred by time $t$, given the covariates $\mathbf{x}_i$.

Furthermore, under the assumption that the event time distribution is continuous, the \textit{cumulative hazard function} $H(t \mid \mathbf{x}_i)$ is related to the survival function via:
\begin{equation}
    S(t \mid \mathbf{x}_i) = \exp(-H(t \mid \mathbf{x}_i)),
\end{equation}
and hence the model output $f(\mathbf{x}_i, t)$ can also be expressed as:
\begin{equation}
    f(\mathbf{x}_i, t) = 1 - \exp(-H(t \mid \mathbf{x}_i)).
\end{equation}
This relation provides an alternative interpretation of the model as implicitly estimating the cumulative hazard function, aggregating the instantaneous risk of failure over time.

Optionally, the instantaneous risk at time $t$ is captured by the \textit{hazard function} $h(t \mid \mathbf{x}_i)$, defined as:
\begin{equation}
    h(t \mid \mathbf{x}_i) := \lim_{\Delta t \to 0} \frac{\Pr(t \leq T < t + \Delta t \mid T \geq t, \mathbf{x}_i)}{\Delta t} = \frac{f_T(t \mid \mathbf{x}_i)}{S(t \mid \mathbf{x}_i)},
\end{equation}
where $f_T(t \mid \mathbf{x}_i)$ denotes the conditional event time density function. The cumulative hazard function is then obtained as the integral of the hazard function:
\begin{equation}
    H(t \mid \mathbf{x}_i) = \int_0^t h(u \mid \mathbf{x}_i) \, du.
\end{equation}
\subsection{Rashomon set}

Given a reference survival model $f_R$ which has the highest prediction performance, a performance metric $\Phi$ (e.g., \textit{concordance index} or \textit{integrated Brier score}), and a tolerance level which is formally defined as the Rashomon parameter $\epsilon > 0$, we define the Rashomon set as:

\begin{equation}
    \mathcal{H}_\epsilon(f_R) := \left\{f \in \mathcal{H} : \Phi(f) \leq \Phi(f_R) + \epsilon \right\}.    
\end{equation}

It captures models whose performance is comparable to the reference within a specified margin. In practice, the choice of $\epsilon$ allows for flexibility when identifying alternative models that may offer similar predictive performance.

\subsection{Measuring model uncertainty}
\label{sec:metric}

In the context of survival analysis, we define a survival risk estimate as \textit{conflicting} if it deviates from the reference survival estimate by at least a specified threshold $\delta \in (0, 1)$. The particular application domain should inform the choice of $\delta$. For instance, a deviation considered critical in industrial prognostics---such as predicting the remaining useful life of a mechanical component or risk estimate of an engine---may differ from what constitutes a significant risk estimate change in other domains like aerospace system monitoring or predictive maintenance in manufacturing. Thus, the notion of a conflicting survival prediction is inherently context-dependent.

\begin{figure}[h!]
    \centering
    \includegraphics[width=\linewidth]{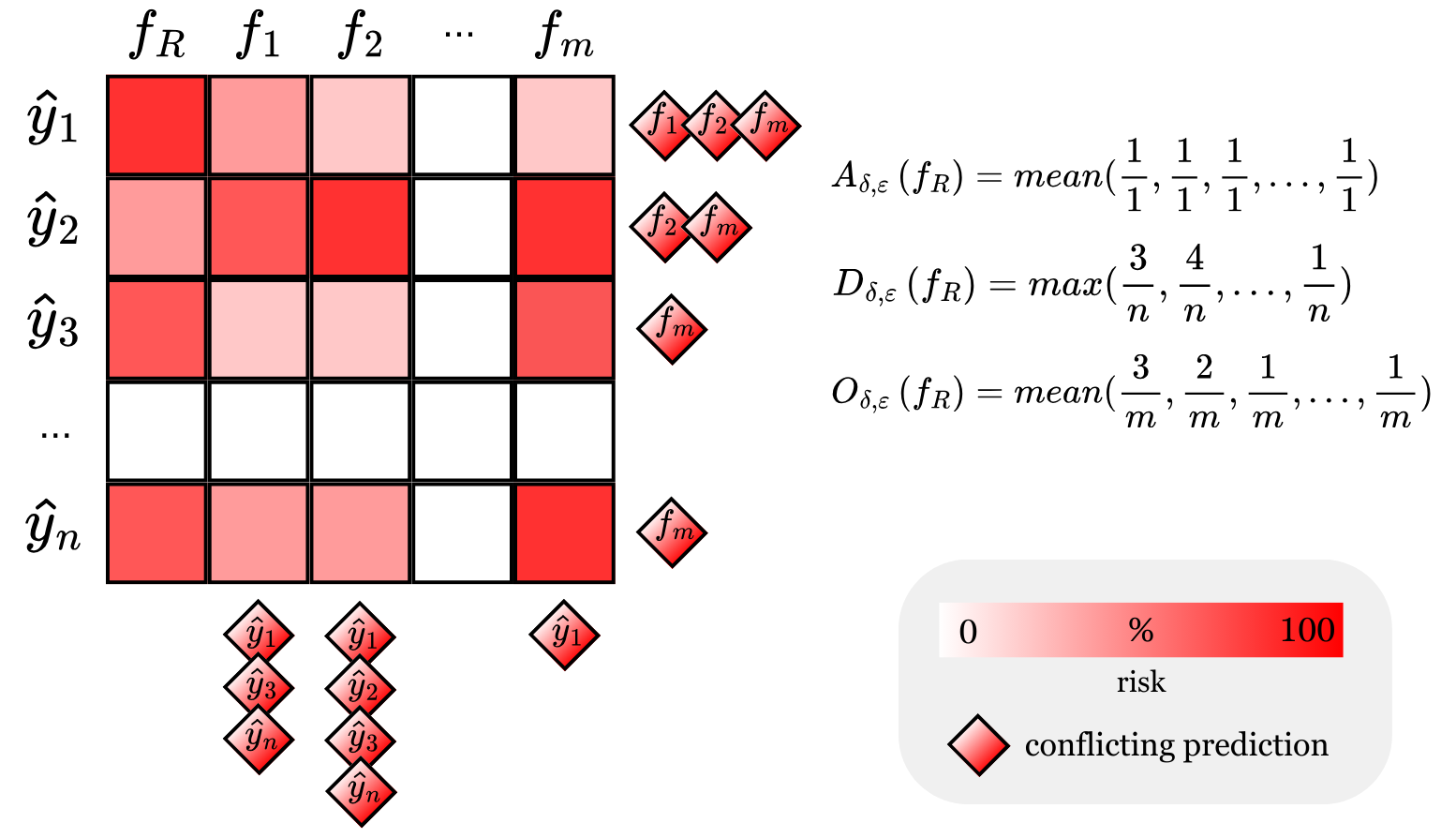}
    \caption{Survival Rashomon cube with size $m \times n$ means that it comprises $m$
models and $n$ observations. The \textit{ambiguity} is the ratio of the models including the conflicting predictions, \textit{discrepancy} is the maximum conflict ratio between the models, and \textit{obscurity} shows the mean conflict ratio across the observations. A conflicting prediction is defined as a risk prediction of a model that deviates from the prediction of the reference model $f_R$ by a $\delta$ difference.}
    \label{fig:rash}
\end{figure}

We modified the predictive multiplicity metrics \textit{ambiguity}, \textit{discrepancy}, and \textit{obscurity} to measure model uncertainty, and introduced them in the following section, inspired by the framework introduced by \citet{marx_et_al_2020}, \citet{watson_et_al_2023}, and \citet{cavus_et_al_2024}, who proposed similar metrics in the context of binary classification. The calculation of metrics is illustrated in Figure~\ref{fig:rash}.

\subsubsection{Ambiguity}

The ambiguity of a survival task over a dataset is the proportion of observations whose reference survival risk estimate changes by at least $\delta$ over the Rashomon set.

\begin{equation}
    A_{\delta, \epsilon}(f_R; \mathcal{D}) := \frac{1}{n} \sum_{i = 1}^n \max_{f \in \mathcal{H}_\epsilon(f_R)} \mathbbm{1}\Big[ \big| f(x_i, t_i) - f_R(x_i, t_i) \big| \geq \delta \Big].
\end{equation}

Relative to the reference survival model $f_R$, ambiguity measures the proportion of individuals for whom the estimated survival risk at time $t_i$ is uncertain by at least $\delta$. High ambiguity indicates more uncertainty in the survival risk predictions over time. Users may use this ambiguity measure along with the reference model's predictions to assess the reliability of risk estimates and guide decision-making processes.

\subsubsection{Discrepancy}

The discrepancy of a survival task over a dataset is the maximum proportion of observations whose survival risk estimates could change by at least $\delta$ by switching the reference survival model $f_R$ with a competing model in the Rashomon set:

\begin{equation}
    D_{\delta, \epsilon}(f_R; \mathcal{D}) := \max_{f \in \mathcal{H}_\epsilon(f_R)} \frac{1}{n} \sum_{i = 1}^n \mathbbm{1} \Big[ \big| f(x_i, t_i) - f_R(x_i, t_i) \big| \geq \delta \Big].    
\end{equation}

Relative to the reference survival model $f_R$, discrepancy reflects the maximum proportion of conflicting survival risk estimates as a result of replacing $f_0$ with another model from the Rashomon set.

\subsubsection{Obscurity}

The obscurity of a survival task over a dataset calculates the average ratio of conflicting survival risk predictions for each observation between the reference survival model $f_R$ and the other models in the Rashomon set:

\begin{equation}
    O_{\delta, \epsilon}(f_R; \mathcal{D}) := \frac{1}{n} \sum_{i=1}^n \frac{1}{|\mathcal{H}_\epsilon(f_R)|} \sum_{f \in \mathcal{H}_\epsilon(f_R)} \mathbbm{1} \Big[\big| f(x_i, t_i) - f_R(x_i, t_i)\big| \geq \delta \Big].
\end{equation}

Relative to the reference survival model $f_R$, obscurity reflects the average proportion of conflicting survival risk estimates as a result of replacing $f_R$ with another model from the Rashomon set.

\section{Experiments}

This section details the datasets and presents the experimental setup of our study. In the following subsections, we first describe the CMAPSS dataset used for modeling engine failures, including its structure and variables. Then, we detail the modeling procedure: we generate a diverse set of survival models by varying hyperparameters.

\subsection{Dataset}

We used the Commercial Modular Aero-Propulsion System Simulation (CMAPSS) dataset \citep{saxena_et_al_2008}, which simulates engine degradation and failures, and includes four distinct subsets reflecting various operating conditions. Because it is recognized as the benchmark dataset for PM applications. The challenges and research perspectives for data are given in \citet{vollert_et_al_2021}. The variables in the CMAPSS dataset and their descriptions and ranges are summarised in Table~\ref{tab:dataset}. The variables marked with — were excluded from the subsets \texttt{FD001} and \texttt{FD003} due to being constant.

\begin{table*}[h!]
\tiny
\centering
\caption{\textbf{Overview of variables in the CMAPSS dataset}}
\label{tab:dataset}\vspace{2mm}
\setlength{\tabcolsep}{7pt}
\renewcommand{\arraystretch}{1.15}
\begin{tabular}{p{1.3cm} p{2cm} p{2cm} p{1.6cm} p{2cm} p{1.6cm}} \toprule
\textbf{Variable} & \textbf{Description} & \texttt{FD001} & \texttt{FD002} & \texttt{FD003} & \texttt{FD004} \\\midrule
\rowcolor{gray!10}
\multicolumn{6}{l}{\textit{Engine \& Cycle Information}} \\\addlinespace[2pt]
\texttt{unit\_number} & Unique engine identifier & 1:100 & 1:260 & 1:100 & 1:260 \\
\texttt{time\_in\_cycles} & Operational cycles since start & 1:362 & 1:378 & 1:525 & 1:378 \\

\addlinespace[6pt]
\rowcolor{gray!10}
\multicolumn{6}{l}{\textit{Operational Settings}} \\\addlinespace[2pt]
\texttt{op\_set\_1} & Operational setting parameter 1 & $[-0.0087,\ 0.0087]$ & $[0,\ 42.008]$ & $[-0.0086,\ 0.0086]$ & $[0,\ 42.008]$ \\
\texttt{op\_set\_2} & Operational setting parameter 2 & $[-6\cdot10^{-4},\ 6\cdot10^{-4}]$ & $[0,\ 0.842]$ & $[-6\cdot10^{-4},\ 7\cdot10^{-4}]$ & $[0,\ 0.842]$ \\
\texttt{op\_set\_3} & Operational setting parameter 3 & — & $[60,\ 100]$ & — & $[60,\ 100]$ \\

\addlinespace[6pt]
\rowcolor{gray!10}
\multicolumn{6}{l}{\textit{Sensor Measurements}} \\\addlinespace[2pt]
\texttt{sensor\_1} & Signal value from the first sensor& — & $[445.00,\ 518.67]$ & — & $[445.00,\ 518.67]$ \\
\texttt{sensor\_2} & Signal value from the second sensor & $[641.21,\ 644.53]$ & $[535.53,\ 644.52]$ & $[640.84,\ 645.11]$ & $[535.53,\ 644.52]$ \\
... & ... & ... & ... & ... & ... \\\addlinespace[6pt]
\texttt{sensor\_21} & Signal value from the twenty-first sensor & $[22.89,\ 23.62]$ & $[6.01,\ 23.59]$ & $[22.87,\ 23.95]$ & $[6.01,\ 23.59]$ \\\bottomrule
\end{tabular}
\end{table*}

\subsection{Modeling}

We first split $80\%$ of the data for training and the remaining $20\%$ for testing. Then, we fixed the censoring time at $t = 250$ to eliminate the censoring sensitivity by taking it as a long period \citep{baskay_et_al_2025}. Following these setups, we generated a model set consisting of Random Survival Forests models \citep{ishwaran_et_al_2008}, which included 22,500 hyperparameter configurations. The detail of hyperparameter configurations is given in Table~\ref{tab:config}. Lastly, we identified the corresponding Rashomon set from the model set using the \textit{Brier score} as the performance metric, with varying Rashomon parameter $\epsilon$ from $0.01$ to $1$. The $\epsilon = 0.01$ means the models, which are performing within $0.01$ of the best performing model, are used as the Rashomon set  $\epsilon = 1$ means the model set is used as the Rashomon set. Some of the Rashomon sets' characteristics in terms of size and performance (\textit{c-index}) for the four subsets of the CMAPSS dataset \texttt{FD001}, \texttt{FD002}, \texttt{FD003}, and \texttt{FD004} under the Rashomon parameters $\epsilon = 0.01, 0.05, 0.10$ are given in Table~\ref{tab:app}. Using the Rashomon set, we quantified predictive uncertainty by applying multiplicity metrics \textit{ambiguity}, \textit{discrepancy}, and \textit{obscurity} adapted for survival tasks, as introduced in Section~\ref{sec:metric}. This procedure was repeated for each dataset, and the resulting findings are presented in the following section.

\begin{table}[h!]
    \centering
    \small
    \caption{\textbf{The detail of hyperparameter configurations}}
    \label{tab:config}\vspace{2mm}
    \begin{tabular}{p{3cm}p{5.6cm}p{3.4cm}}\toprule
        Hyperparameter      & Description   & Values \\\midrule
        \texttt{ntree}      & Number of trees to grow in the forest & $\{100, 300, ..., 1900\}$\\
        \texttt{mtry}       & Number of variables randomly selected at each split & $\{1, 3, ..., 9\}$\\
        \texttt{nodesize}   & Minimum number of samples required to split an internal node & $\{5, 25, ..., 85\}$\\
        \texttt{nodedepth}  & Maximum depth of a tree & $\{5, 25, ..., 85\}$\\
        \texttt{splitrule}  & Splitting criterion used to decide the best split at each node & $\{$\texttt{logrank}, \texttt{logrankscore}, \texttt{bs.gradient}$\}$\\
        \texttt{nsplit}     & Number of random split points considered for each candidate split variable & $\{5, 7, ..., 15\}$\\\bottomrule
    \end{tabular}
\end{table}

\section{Results}

Table~\ref{tab:res} presents the values of the \textit{ambiguity}, \textit{discrepancy}, and \textit{obscurity} metrics calculated under specific values of the Rashomon parameter $\epsilon$, defines how close a model's performance must be to the reference model to be included in the Rashomon set, and the prediction deviation threshold $\delta$ for four different datasets: \texttt{FD001}, \texttt{FD002}, \texttt{FD003}, and \texttt{FD004}. Here, $\delta$ sets the threshold for how different a model's prediction must be from that of the reference model for a given observation to be considered conflicting.

At the smallest considered Rashomon parameter, $\epsilon = 0.01$, the \texttt{FD003} dataset exhibits notably high values of both \textit{ambiguity} and \textit{discrepancy}---reaching $1$ at certain $\delta$ thresholds. This indicates that models with nearly the same performance cannot agree on the prediction of any observation. A similar, though less pronounced, pattern is observed for \texttt{FD004}. On the other hand, the \texttt{FD001} dataset exhibits \textit{ambiguity} only at $\delta = 0.01$, and does not show it at higher $\delta$ values. This suggests that, within a tightly defined Rashomon set, \texttt{FD001} predictions show a deviation of 0.01 from the reference model, but no deviation above 0.05. Moreover, the Rashomon set size under this narrow tolerance is extremely limited (e.g., a single model for \texttt{FD002}), supporting the low uncertainty observed.

As $\epsilon$ increases to $0.05$, a broader range of models---those with slightly worse performance than the reference model---are included in the Rashomon set. Under this condition, \textit{ambiguity} and \textit{discrepancy} increase across all datasets except \texttt{FD003}. For example, in datasets other than \texttt{FD001}, \textit{ambiguity} values range between $0.85$ and $1$, indicating that many of the additional models produce differing predictions despite their acceptable performance. This pattern is reflected in the \textit{obscurity} metric: for instance, \texttt{FD002} reaches an \textit{obscurity} value of $0.8875$ at $\delta = 0.01$, meaning that a large proportion of individual predictions diverge from those of the reference model by more than the threshold. The increase in Rashomon set size (e.g., $4135$ models for \texttt{FD004}) further confirms that model diversity leads to greater uncertainty and reinforces the importance of model selection even among similarly performing alternatives.

At the highest considered Rashomon parameter, $\epsilon = 0.10$, the Rashomon set includes a wide range of models with performance close---but not identical---to the reference model. Here, both \textit{ambiguity} and \textit{discrepancy} reach or approach $1$ across most datasets and $\delta$ values. This means that predictions vary significantly across almost all models in the Rashomon set, depending on which one is used. \textit{Obscurity} values follow a similar trend. For instance, \texttt{FD004} shows an \textit{obscurity} of 0.9028 at $\delta = 0.01$. However, as $\delta$ increases, a relative decline in both \textit{ambiguity} and \textit{discrepancy} is observed. This implies that, while small prediction differences are widespread, fewer observations exceed larger deviation thresholds. Nonetheless, the substantial number of models in the Rashomon set at this level (e.g., 11328 for \texttt{FD002}) maintains a high level of uncertainty.

\begin{table}
    \centering
    \footnotesize
    \caption{The values of \textit{ambiguity} $A_{\epsilon, \delta}$, \textit{discrepancy} $D_{\epsilon, \delta}$, and \textit{obscurity} $O_{\epsilon, \delta}$ for the four subsets of the CMAPSS dataset \texttt{FD001}, \texttt{FD002}, \texttt{FD003}, and \texttt{FD004} under the Rashomon parameters $\epsilon = 0.01, 0.05, 0.10$ and the thresholds $\delta = 0.01, 0.05, 0.10$. Here, $\delta$ defines the threshold for how much a model’s prediction must differ from that of the reference model for a given observation to be considered conflicting.}
    \label{tab:res} \vspace{2mm}
    \begin{tabular}{ccccccc}\toprule
        $\epsilon$  & $\delta$  & metric                & \texttt{FD001} & \texttt{FD002} & \texttt{FD003} & \texttt{FD004} \\\midrule
                    &           & $A_{\epsilon, \delta}$& 0.9            & -              & 1              & 0.66 \\
                    & 0.01      & $D_{\epsilon, \delta}$& 0.7            & -              & 1              & 0.66 \\
                    &           & $O_{\epsilon, \delta}$& 0.325          & -              & 0.9875         & 0.66 \\\cline{2-7}
                    &           & $A_{\epsilon, \delta}$& 0              & -              & 0.9            & 0.1 \\
        0.01        & 0.05      & $D_{\epsilon, \delta}$& 0              & -              & 0.9            & 0.1 \\
                    &           & $O_{\epsilon, \delta}$& 0              & -              & 0.85           & 0.1 \\\cline{2-7}
                    &           & $A_{\epsilon, \delta}$& 0              & -              & 0.8            & 0.02 \\
                    & 0.10      & $D_{\epsilon, \delta}$& 0              & -              & 0.7            & 0.02 \\
                    &           & $O_{\epsilon, \delta}$& 0              & -              & 0.575          & 0.02 \\\midrule
                    
                    &           & $A_{\epsilon, \delta}$& 1              & 1              & 1              & 1 \\
                    & 0.01      & $D_{\epsilon, \delta}$& 1              & 1              & 1              & 1 \\
                    &           & $O_{\epsilon, \delta}$& 0.4751         & 0.8875         & 0.8789         & 0.86 \\\cline{2-7}
                    &           & $A_{\epsilon, \delta}$& 0.85           & 1              & 1              & 1 \\
        0.05        & 0.05      & $D_{\epsilon, \delta}$& 0.55           & 0.8653         & 0.95           & 0.86 \\
                    &           & $O_{\epsilon, \delta}$& 0.0157         & 0.4894         & 0.6989         & 0.5188\\\cline{2-7}
                    &           & $A_{\epsilon, \delta}$& 0.1            & 1              & 0.95           & 0.94 \\
                    & 0.10      & $D_{\epsilon, \delta}$& 0.05           & 0.6538         & 0.75           & 0.7 \\
                    &           & $O_{\epsilon, \delta}$& 0.0004         & 0.2256         & 0.4120         & 0.2219\\\midrule
                    
                    &           & $A_{\epsilon, \delta}$& 1              & 1              & 1              & 1 \\
                    & 0.01      & $D_{\epsilon, \delta}$& 1              & 1              & 1              & 1 \\
                    &           & $O_{\epsilon, \delta}$& 0.7501         & 0.9302         & 0.8576         & 0.9028 \\\cline{2-7}
                    &           & $A_{\epsilon, \delta}$& 1              & 1              & 1              & 1 \\
        0.10        & 0.05      & $D_{\epsilon, \delta}$& 0.6            & 0.9038         & 0.95           & 0.92 \\
                    &           & $O_{\epsilon, \delta}$& 0.1169         & 0.6667         & 0.6788         & 0.6475 \\\cline{2-7}
                    &           & $A_{\epsilon, \delta}$& 0.3            & 1              & 0.95           & 0.96 \\
                    & 0.10      & $D_{\epsilon, \delta}$& 0.2            & 0.6923         & 0.75           & 0.74 \\
                    &           & $O_{\epsilon, \delta}$& 0.0037         & 0.4171         & 0.4274         & 0.4005\\\bottomrule
                    
    \end{tabular}
\end{table}

Overall, model uncertainty is limited when $\epsilon$ is small but increases markedly as $\epsilon$ grows. This reinforces the utility of the Rashomon set as a framework for analyzing model reliability. Among the datasets, \texttt{FD001} demonstrates the lowest level of uncertainty, whereas \texttt{FD003} and \texttt{FD004} display higher variability. These differences likely reflect inherent characteristics of the datasets, such as internal complexity, signal-to-noise ratio, or alignment with the model structure. 

\begin{figure}
    \centering
    \includegraphics[width=0.9\linewidth]{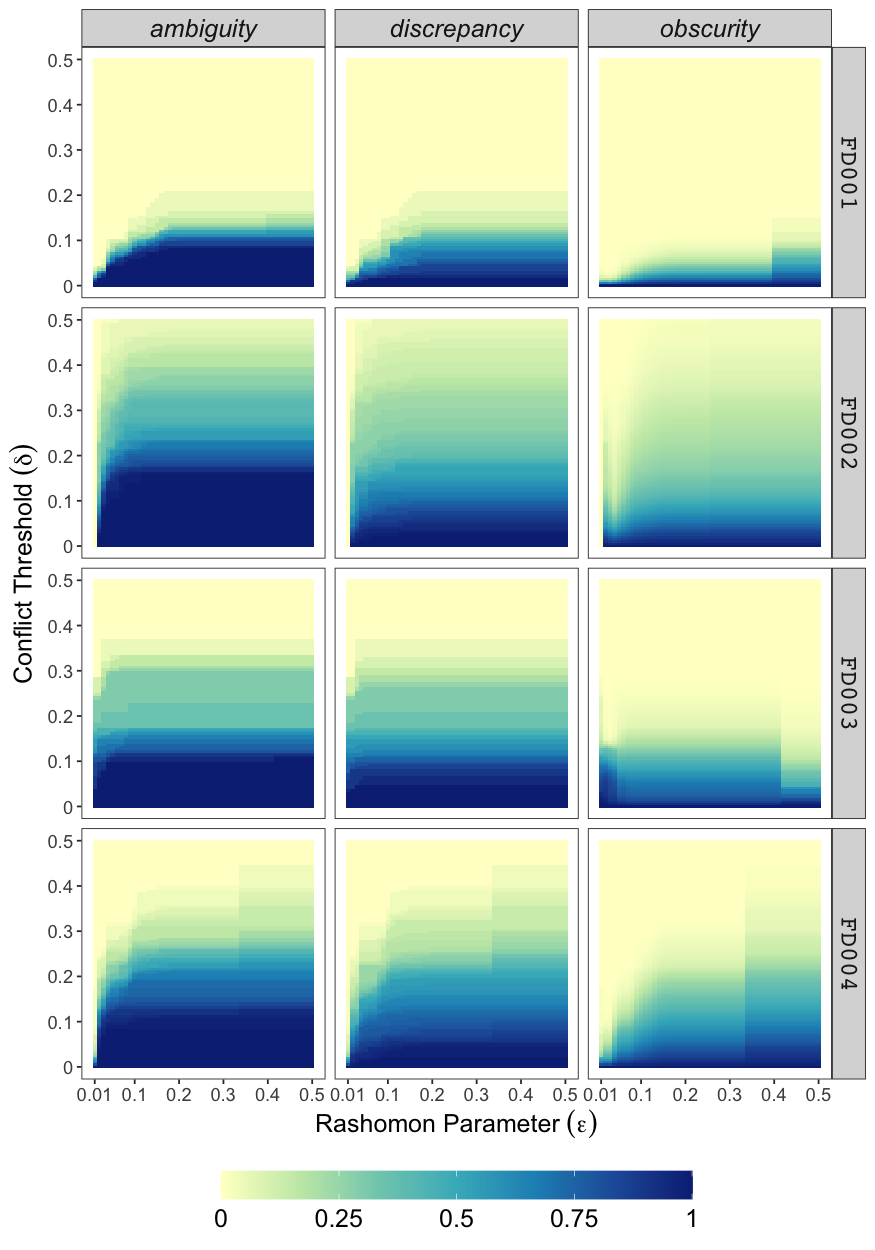}
    \caption{The values of predictive multiplicity metrics across various Rashomon parameters $\epsilon$ and conflict thresholds $\delta$ for the four subsets of the CMAPSS dataset. The color gradient represents the severity of multiplicity. $\delta$ sets the threshold for how much a model's prediction must differ from the reference to be considered conflicting.}
    \label{fig:res}
\end{figure}

However, these results are limited to a few values of $\epsilon$ and $\delta$. We extended the results for $\epsilon$ and $\delta$ taken between $0$ and $0.5$. In Figure~\ref{fig:res}, we present the values of predictive multiplicity metrics across all combinations of the Rashomon parameter ($\epsilon$) and error tolerance threshold ($\delta$) for four subsets of the CMAPSS dataset. The intensity of the colors corresponds to the metric value on the $[0, 1]$ scale, with darker tones indicating higher values and lighter tones indicating lower values. 

The \textit{ambiguity} metric measures the proportion of instances whose estimated survival risk under the reference model $f_R$ changes by at least $\delta$ under at least one model in the Rashomon set. A high \textit{ambiguity} value indicates greater uncertainty in individual risk predictions over time and suggests caution in decision-making. As seen in the figure, for \texttt{FD001}, \textit{ambiguity} increases with $\epsilon$ only at low values of $\delta$, suggesting that substantial divergence from the reference model occurs only under lenient conflict thresholds. For \texttt{FD002}, \texttt{FD003}, and \texttt{FD004}, \textit{ambiguity} generally increases as $\delta$ decreases, and becomes more pronounced as $\epsilon$ increases. This suggests that, in these datasets, a higher proportion of instances are susceptible to risk estimate variation across the Rashomon set.

The \textit{discrepancy} metric reflects the maximum proportion of instances whose risk predictions could change by at least $\delta$ when replacing the reference model with any other model in the Rashomon set. As an indicator of worst-case disagreement, high discrepancy values imply that substantial deviations in model outputs are possible. For \texttt{FD001} and \texttt{FD002}, discrepancy patterns closely follow those of \textit{ambiguity}, rising at low $\delta$ and high $\epsilon$, indicating the presence of competing models with potentially drastic outcome differences. In \texttt{FD003} and \texttt{FD004}, the discrepancy is more evenly distributed and decreases more clearly with increasing $\delta$, implying that even the most divergent model in the Rashomon set remains relatively consistent with the reference model.

The \textit{obscurity} metric captures the average proportion of conflicting survival risk predictions across the entire Rashomon set for each individual. Unlike discrepancy, which reflects worst-case deviation, \textit{obscurity} summarizes the general disagreement level within the Rashomon set. In \texttt{FD001}, \textit{obscurity} is high only at small values of $\delta$ and $\epsilon$, then declines sharply with increasing $\epsilon$. This suggests that substantial conflict exists within small Rashomon sets but becomes diluted in larger ones. \texttt{FD002} follows a similar pattern, with peak obscurity at low $\epsilon$ and rapid decay as the set size increases, indicating that smaller subsets of models may strongly diverge from the reference. \texttt{FD003} shows relatively low \textit{obscurity} overall, reflecting consistent risk estimates across the Rashomon set. FD004 exhibits high obscurity at small $\epsilon$ and $\delta$ values, which gradually fades as $\epsilon$ increases, implying that while numerous conflicting models exist in tighter Rashomon sets, these conflicts are not sustained when the set is expanded.

In summary, the \textit{ambiguity} and \textit{discrepancy} metrics generally increase with smaller $\delta$ and larger $\epsilon$, indicating greater potential for conflicting predictions. In contrast, \textit{obscurity} often peaks at low $\epsilon$ and then declines, showing that disagreement is most pronounced among smaller, tighter Rashomon sets. Taken together, these metrics provide complementary insights into model uncertainty and reliability by assessing not only the extent of divergence from a reference model but also the consistency across plausible alternative models.
\section{Conclusions}

In this paper, we adapted a novel framework, predictive multiplicity, to survival models, specifically within the domain of predictive maintenance. By formalizing three complementary metrics—\textit{ambiguity}, \textit{discrepancy}, and \textit{obscurity}—over the Rashomon set, we captured distinct aspects of model disagreement under varying Rashomon parameters and conflict thresholds. This approach aims to improve process health management by better characterizing uncertainty in predictions related to equipment degradation and potential failure.

Our empirical evaluation on the CMAPSS datasets reveals considerable variation in multiplicity depending on the combination of performance tolerance and conflict thresholds. At the narrowest Rashomon parameter, we observed that the \texttt{FD003} dataset exhibits the highest levels of disagreement—\textit{ambiguity} and \textit{discrepancy} both reach their maximum values for the smallest conflict threshold—indicating that even nearly indistinguishable models in terms of performance can produce fully divergent predictions. A similar, although milder, trend is visible in \texttt{FD004}. In contrast, \texttt{FD001} shows only localized multiplicity (e.g., \textit{ambiguity} of nearly one at the smallest conflict threshold), and some datasets (such as \texttt{FD002}) contain no Rashomon set members at all for low Rashomon parameters, underscoring their low model uncertainty under severe model similarity.

As the Rashomon parameter increases, broader Rashomon sets emerge, encompassing thousands of near-optimal models, for instance, more than eleven thousand models in \texttt{FD002} at the higher Rashomon parameters. This expansion correlates with substantial increases in both \textit{ambiguity} and \textit{discrepancy}, which often reach or approach their maximum values across multiple datasets. \textit{Obscurity} follows similar trends in \texttt{FD004}, for example, it rises to a value slightly above $0.9$ at the smallest conflict threshold. However, for larger thresholds, these metrics decline, reflecting that while small deviations in predictions are pervasive, extreme disagreements are more limited.

Despite the strengths of this framework, several limitations remain. First, our analysis focuses exclusively on random survival forests models applied to the CMAPSS datasets. Extending this approach to other model classes (e.g., deep survival networks) and domains (e.g., healthcare, finance) is a promising direction for future work. Second, our multiplicity metrics depend on choices of performance metric (e.g., \textit{integrated Brier score}) and conflict threshold; alternative definitions may yield different Rashomon set structures and multiplicity profiles. Third, our metrics operate solely based on survival risk estimates, though they can be generalized to other types of survival outputs such as time-to-event predictions.

Looking ahead, future research may explore strategies to mitigate or manage predictive multiplicity, such as model aggregation via ensembles, robust selection within the Rashomon set, or decision-making frameworks that incorporate \textit{ambiguity}, \textit{discrepancy}, and \textit{obscurity} explicitly. Integrating these multiplicity-aware tools into cost-sensitive and risk-averse maintenance planning pipelines could lead to the reliable and transparent deployment of predictive models in process health management, where accurate identification of equipment degradation and timely intervention before system failure are critical. Moreover, multiplicity-aware evaluation could aid in validating and building confidence in deployed models by revealing the full spectrum of plausible predictions. Beyond predictive maintenance, this framework could also be extended to high-stakes domains such as healthcare, where incorporating predictive multiplicity into survival analysis may support more reliable, individualized decision-making under uncertainty and enhance trust in risk-based clinical assessments. Additionally, it is worth considering the potential contributions that uncertainty quantification could bring to active learning \citep{nguyen_et_al_2022}. By quantifying the model uncertainty, active learning strategies could prioritize the most uncertain instances for labeling, leading to more efficient learning processes and improved model performance, especially in scenarios where labeled data is scarce.

\section*{Supplemental Materials}

The materials for reproducing the experiments and benchmark data sets can be found in the repository: \url{https://github.com/mcavs/survival_predictive_multiplicity_paper}.

\section*{Declaration of generative AI and AI-assisted technologies in the writing process}

While preparing this paper, the author used ChatGPT 4.0 for grammatical correction. After using this tool, the author reviewed and edited the content as needed and took full responsibility for the publication's content.

\appendix
\section{Rashomon sets' characteristics}
\label{app1}

\setcounter{table}{0}
\begin{table}[h!]
    \centering
    \small
    \caption{Rashomon sets' characteristics in terms of size and performance (\textit{c-index}) for the four subsets of the CMAPSS dataset \texttt{FD001}, \texttt{FD002}, \texttt{FD003}, and \texttt{FD004} under the Rashomon parameters $\epsilon = 0.01, 0.05, 0.10$}
    \label{tab:app}\vspace{2mm}
    \begin{tabular}{lccccc}\toprule
                            &$\epsilon$ & \texttt{FD001}    & \texttt{FD002}    & \texttt{FD003}    & \texttt{FD004}    \\\midrule
                            &$0.01$     & 7                 & 1                 & 5                 & 2                 \\
    Rashomon set size       &$0.05$     & 714               & 1755              & 2984              & 4135              \\
                            &$0.10$     & 3396              & 11328             & 9000              & 7829              \\\midrule
                            &$0.01$     & [0.88, 0.89]  & [0.74, 0.75]  & [0.90, 0.91]  & [0.82, 0.83]  \\
    Performance             &$0.05$     & [0.84, 0.89]  & [0.70, 0.75]  & [0.86, 0.91]  & [0.78, 0.83]  \\
                            &$0.10$     & [0.79, 0.89]  & [0.65, 0.75]  & [0.81, 0.91]  & [0.73, 0.83]  \\\bottomrule
    \end{tabular}
\end{table}


\end{document}